\documentclass[lettersize,journal]{IEEEtran}
\usepackage{amsmath,amsfonts}
\usepackage{algorithmic}
\usepackage{algorithm}
\usepackage{array}
\usepackage[caption=false,font=normalsize,labelfont=sf,textfont=sf]{subfig}
\usepackage{textcomp}
\usepackage{stfloats}
\usepackage{url}
\usepackage{verbatim}
\usepackage{graphicx}
\usepackage{cite}
\usepackage{bigstrut,multirow,rotating,booktabs}
\usepackage{booktabs} 
\usepackage{tabularx} 
\usepackage{caption} 
\usepackage{stfloats}
\usepackage[colorlinks=true,linkcolor=blue,urlcolor=blue,citecolor=blue]{hyperref}
\usepackage{algorithm}
\usepackage{graphicx}
\usepackage{float}
\DeclareMathOperator*{\argmin}{arg\,min}
\usepackage[switch]{lineno}
\begin{document}

\title{ YOCO: You Only Calibrate Once for Accurate\\Extrinsic  Parameter in LiDAR-Camera Systems}

\author{Tianle Zeng, Dengke He, Feifan Yan, Meixi He
\thanks{
Tianle Zeng is with the State Key Labortaory for Fine Exploration and Intelligent Development of Coal Resources, China University of Mining and Technology (Beijing), Beijing 100083, China, and also with the College of Geoscience and Surveying Engineering, China University of Mining and Technology (Beijing), Beijing 100083, China (e-mail: zqt2200202081@student.cumtb.edu.cn).\\
Dengke He is with the State Key Labortaory for Fine Exploration and Intelligent Development of Coal Resources, China University of Mining and Technology (Beijing), Beijing 100083, China, and also with the College of Geoscience and Surveying Engineering, China University of Mining and Technology (Beijing), Beijing 100083, China (e-mail: he\_dengke@126.com).\\
Feifan Yan is with the State Key Labortaory for Fine Exploration and Intelligent Development of Coal Resources, China University of Mining and Technology (Beijing), Beijing 100083, China, and also with the College of Geoscience and Surveying Engineering, China University of Mining and Technology (Beijing), Beijing 100083, China (e-mail:  yabiyff@163.com).\\
Meixi He is with the State Key Labortaory for Fine Exploration and Intelligent Development of Coal Resources, China University of Mining and Technology (Beijing), Beijing 100083, China, and also with the College of Geoscience and Surveying Engineering, China University of Mining and Technology (Beijing), Beijing 100083, China (e-mail: zqt2200202050@student.cumtb.edu.cn).
}}

\markboth{IEEE TRANSACTIONS ON INSTRUMENTATION AND MEASUREMENT,~Vol.~x, No.~x, x~2024}%
{Shell \MakeLowercase{\textit{et al.}}: A Sample Article Using IEEEtran.cls for IEEE Journals}


\maketitle

\begin{abstract}
In a multi-sensor fusion system composed of cameras and LiDAR, precise extrinsic calibration contributes to the system's long-term stability and accurate perception of the environment. However, methods based on extracting and registering corresponding points still face challenges in terms of automation and precision. This paper proposes a novel fully automatic extrinsic calibration method for LiDAR-camera systems that circumvents the need for corresponding point registration. In our approach, a novel algorithm to extract required LiDAR correspondence point is proposed. This method can effectively filter out irrelevant points by computing the orientation of plane point clouds and extracting points by applying distance- and density-based thresholds. We avoid the need for corresponding point registration by introducing extrinsic parameters between the LiDAR and camera into the projection of extracted points and constructing co-planar constraints. These parameters are then optimized to solve for the extrinsic. We validated our method across multiple sets of LiDAR-camera systems. In synthetic experiments, our method demonstrates superior performance compared to current calibration techniques. Real-world data experiments further confirm the precision and robustness of the proposed algorithm, with average rotation and translation calibration errors between LiDAR and camera of less than 0.05° and 0.015m, respectively. This method enables automatic and accurate extrinsic calibration in a single one step, emphasizing the potential of calibration algorithms beyond using corresponding point registration to enhance the automation and precision of LiDAR-camera system calibration.

\end{abstract}


\section{Introduction}
\IEEEPARstart{S}{ENSOR}  fusion has been widely discussed in the robotics and computer vision fields. The complementary characteristics between sensors can capitalize on the advantages of different sensors and compensate for the shortcomings, which makes the multi-sensor system have higher accuracy and robustness than the single-sensor system. Cameras provide a cost-effective solution for perception systems, offering high-resolution color images with deep-learning capabilities for tasks like object detection and semantic segmentation \cite{8953311}, \cite{zhang2018occlusion}. However, they lack direct depth measurement and are less efficient in low-light environments. 3D Light Detection And Ranging(LiDAR) sensors generate accurate point clouds independently of ambient light, overcoming these limitations. Although LiDAR point clouds can be sparse with lower refresh rates, combining LiDAR and camera measurements compensates for their respective shortcomings. The success of sensor fusion critically depends on the precise calibration of extrinsic parameters. These parameters are primarily computed through the extraction and registration of corresponding points between the camera and LiDAR sensor. Therefore, the essence of LiDAR-camera system calibration lies in establishing and registering corresponding relationships among co-observable points. 

However, both establishing corresponding relationships and registering correspondence points are highly challenging tasks. In establishing corresponding relationships, main approaches involve identifying corresponding features between cameras and LiDAR sensors. Previous studies have concentrated on geometric features as well as artificial features \cite{humeau2019texture}. Typically, detecting features in 2D images is straightforward, but identifying and extracting corresponding 3D feature points in the LiDAR point cloud presents a more complex challenge. This is because 2D images contain rich semantic and color information, while point clouds often only consist of coordinates in 3D space. These sparsely distributed point clouds make it difficult to find suitable features. To solve this problem, a general approach is to select corresponding points manually \cite{park2014calibration}. Once feature points, such as corner points, are detected in images, manual selection and extraction of matching points in LiDAR point clouds are performed. For manually selected correspondence points, point cloud registration is relatively straightforward, with common methods including point-to-point, point-to-line, line-to-line, and line-to-plane registration\cite{zheng2013revisiting}, \cite{huang2021comprehensive}, \cite{fitzgibbon2003robust}, \cite{aoki2019pointnetlk}. Due to the relative accuracy of manually selected correspondence points, high-precision extrinsic parameters can be obtained through point cloud registration. However, manual selection of correspondence points entails significant human intervention, leading to a complex, cumbersome, and time-consuming calibration process.

In recent years, automated methods for selecting and extracting corresponding points have gained prominence in the field, aimed at reducing manual intervention and enhancing calibration efficiency \cite{cheng2018registration}. These methods primarily involve identifying common geometric features and fitting various geometric shapes of different dimensions, such as 1-D lines, 2-D planes, and 3-D spheres, in both images and LiDAR point clouds \cite{9599702,10301475}. However, while the automated extraction method of correspondence points minimizes human intervention, it brings challenges to subsequent point cloud registration, particularly in terms of time and precision.
The time-related challenge arises due to the limited availability of point clouds meeting the feature point extraction requirements, necessitating the collection of a large amount of input data. Consequently, processing this large volume of data prolongs the algorithm execution times. Moreover, learning-based methods require extensive data collection and prolonged training time to achieve satisfactory results.
Additionally, the precision-related challenge stems from the fact that automatically extracted corresponding feature points may not guarantee complete accuracy. This can lead to misalignment issues that affect calibration accuracy. Furthermore, the extensive demand for input data in the point cloud registration process further leads to more misaligned points being used for registration, ultimately impacting accuracy. Although several methods \cite{yang2020teaser,li2020evaluation} have been proposed to optimize the processes of point cloud extraction and registration, they still cannot fundamentally resolve the issue.

Motivated by these challenges, this paper introduces a novel calibration method that automatically extracts corresponding points and circumvents the point cloud registration process through innovative extrinsic calculation, addressing challenges in time and precision. This approach minimizes errors caused by mismatched corresponding points, maximizing point cloud information utilization, reducing calibration input data requirements, and enhancing precision and speed. The proposed method is straightforward to implement, requiring minimal data for high-precision calibration without manual intervention. Extensive simulations and real-world experiments on various LiDAR-camera setups, including publicly available datasets, demonstrate the versatility and effectiveness of our approach. The main contributions of this study are summarized as follows:

1)We propose a novel method for extrinsic parameter calculation that eliminates the necessity of registering correspondence points between cameras and LiDAR sensors, this offers a fresh perspective and technical solution to the conventional calibration process.

2)We introduce a novel plane extraction method for detecting and extracting correspondence plane point clouds, which provides valuable insight for plane extraction scenarios involving prior knowledge of spatial geometry.

3)We present a fully automatic calibration pipeline between LiDAR and camera systems. This process has a simple requirement for calibration data collection and can be applied for the offline calibration of various LiDAR and camera configurations, The source code will be available at \url{https://github.com/louiszengCN/lidar_camera_auto_calibration}.

The remainder of this paper is organized as follows. We briefly survey the field of LiDAR-camera system calibration in Section \ref{sec:relativework}. An overview of the proposed method is given in Section \ref{sec:overview}. After that, the proposed methodology is presented in Sections \ref{sec:methodology}. In Section \ref{sec:experiment} we evaluate our method with simulation and real-world experiments. Finally, we conclude our work in Section \ref{sec:conclusion}.

\section{Relative work}\label{sec:relativework}
Extrinsic calibration techniques are commonly classified into two main categories: target-based methods and target-less methods. Each of these approaches has been thoroughly researched and implemented in practice, with a detailed examination of their respective strengths and drawbacks.
\subsection{Target-Based Methods}
The target-based method has received extensivee research attention, covering manual, semi-automatic, fully automatic methodologies. It is extensively utilized in the calibration of cameras and LiDAR systems.  Zhang et al.  \cite{zhang2004extrinsic} use a checkerboard as a planar geometry feature and then extract this feature both in camera and laser point to calibrate the lidar and camera. Zhao et al.  \cite{zhao2007efficient} project points in laser and cameras to a special coordinate system to calculate the ridge body transformation matrix between laser and camera. Gong et al.  \cite{gong20133d} extract trihedron corner points from laser and camera data and calibrate the extrinsic parameter by matching the corner points. However, these approaches necessitate manual extraction of feature points in the LiDAR point cloud. Methods requiring manual involvement have a high level of calibration precision, but it often has specific data acquisition requirements, limited applicability, and complexities in the calibration process, demanding significant time and cost investments.

There are methods that do not require manual intervention and can automatically extract LiDAR point cloud data for calibration. Geiger et al.  \cite{geiger2012automatic} employ Principal Component Analysis (PCA) to compute normal vectors for each data point within the point cloud. Subsequently, they utilize a greedy region-growing process to segment the chessboard plane. Pandey et al.  \cite{pandey2010extrinsic} use a checkerboard pattern as the co-observable features in both camera and laser data and apply Random Sampling Consensus (RANSAC) plane fitting algorithm to extract point clouds of plane features in LiDAR data. Toth et al.  \cite{toth2020automatic} use Spheres as a calibration target, employing the RANSAC algorithm to detect the inlier points and formulating a least-squares optimization problem based on the radius constraint to extract the sphere point cloud from the LiDAR data. 

Relative to targetless methods, target-based calibration is typically characterized by higher precision and robustness, with the added advantage of traceable calibration errors. This makes it particularly well-suited for achieving precise calibration within controlled settings. Consequently, the method proposed in this paper employs a target-based approach, utilizing a printed checkerboard as the calibration target.

\subsection{Targetless-Based Methods}
The targetless method  \cite{liu2022targetless} endeavors to identify natural patterns, predominantly lines or orthogonal features, within the scene. Subsequently, it formulates these patterns using geometric constraints to deduce the extrinsic parameters. This methodology encompasses two primary approaches: the static-based method and motion-based methods.

For the static-based methodologies, it is assumed that the camera remains stationary or moves slowly relative to the scene. Key features such as lines, corners, or other discernible patterns are detected within the scene, and their geometric attributes are leveraged to estimate the camera's pose. Techniques such as structure-from-motion (SfM) or simultaneous localization and mapping (SLAM) can be applied within this framework. Nie et al.  \cite{nie2022automatic} presented a novel approach for dual LiDAR calibration based on adaptive surface normal estimation. Zhang et al.  \cite{zhang2023overlap} introduced a method that eliminates the need for overlapping field of view (FOV) between LiDAR and camera. Instead, external parameters are obtained from odometry estimation conducted independently by the sensor. Additionally, recent advancements in deep learning techniques have been utilized for targetless multisensor calibration. 
Ye et al.  \cite{ye2021keypoint} proposed a network for 2D-3D pose estimation based on keypoints. This approach embeds an optimizer based on geometric constraints into an end-to-end network and employs a trainable point weighting layer to extract sparse corresponding points for calibration. Zhu et al. \cite{zhu2023robust} trained a cross-modal graph neural network (GNN) for the purpose of calibrating a LiDAR-Camera system. They utilize PointNet and PointNet++ models, which are integral for extracting feature points from point cloud data. Motion-based methods exploit the motion of the camera or sensor to estimate its pose. By analyzing the apparent motion of features in the scene over time, the camera's trajectory and orientation can be inferred. Optical flow, visual odometry, or feature tracking algorithms are commonly used in motion-based methods.

Nevertheless, these methods encounter specific challenges. Firstly, ensuring the generalization ability of calibration results across various scenarios proves difficult due to the variability and diversity of data within different scenes, which can influence calibration outcomes. Secondly, a deficiency exists in real-time evaluation techniques to adequately assess calibration results. This deficiency is crucial for ensuring the quality and reliability of sensor fusion systems in practical field applications.

\section{OVERVIEW}\label{sec:overview}
\begin{figure}
    \centering
    \includegraphics[width=1\linewidth]{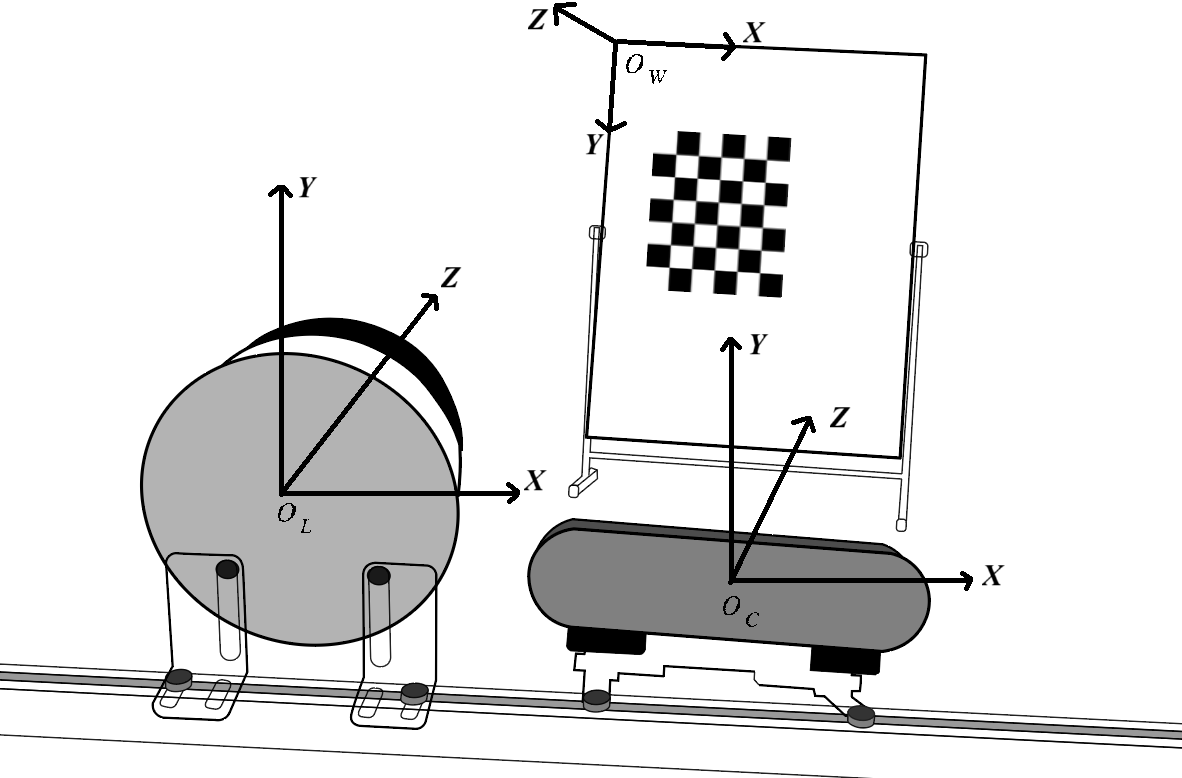}
    \caption{A LiDAR-camera system, \( O_{\text{L}} \) and \( O_{\text{C}} \) denotes the coordinate system of LiDAR and camera. The \(xoy\) plane of word coordinate system \(O_{\text{W}}\) is attached to the plane of checkerboard in this paper.}
    \label{fig:overview}
\end{figure}
As depicted in Fig.\ref{fig:overview}, this method utilizes a checkerboard as the calibration target. A LiDAR and a camera are mounted on a carrier. Their coordinate systems are denoted as \( O_{\text{L}} \) and \( O_{\text{C}} \) respectively. We define the \( xoy \) plane of the world coordinate system as \( O_{\text{w}} \), which is aligned with the checkerboard plane. The objective is to determine the rigid transformation relationship between the LiDAR and the camera, expressed as:
\begin{equation}
\label{deqn_ex1}
 P_{\text{C}}^i = R_{\text{CL}} \cdot P_{\text{L}}^i + t_{\text{CL}} 
\end{equation}

Here, \( P_{\text{C}}^i \) and \( P_{\text{L}}^i \) represent a pair of point clouds captured by the LiDAR and the camera, respectively. \( R_{\text{CL}} \) denotes the rotation matrix, and \( t_{\text{CL}} \) is the translation vector. \( R_{\text{CL}} \) and \( t_{\text{CL}} \) constitute the final output of the proposed calibration method.

The method proposed in this paper enables calibration in both simple and complex indoor and outdoor environments. While a clean environment facilitates algorithmic performance, the algorithm remains effective in complex settings (as discussed further in Section \ref{sec:methodology}). The calibration process is straightforward: a large chessboard pattern is printed for use as the checkerboard, ensuring proper alignment with the camera during calibration to ensure the complete visibility of the pattern within the camera's field of view. These conditions are relatively easy to meet, especially for offline calibration. Following the simultaneous collection of checkerboard data with both LiDAR and camera systems, the algorithm automatically conducts extrinsic calibration, yielding the required \( R_{\text{CL}} \) and \( t_{\text{CL}} \) parameters without the need for manual intervention. The entire calibration process is efficient and straightforward, boasting high precision (as extensively discussed in Section \ref{sec:experiment}).

\begin{figure*}
    \centering
    \includegraphics[width=1\linewidth]{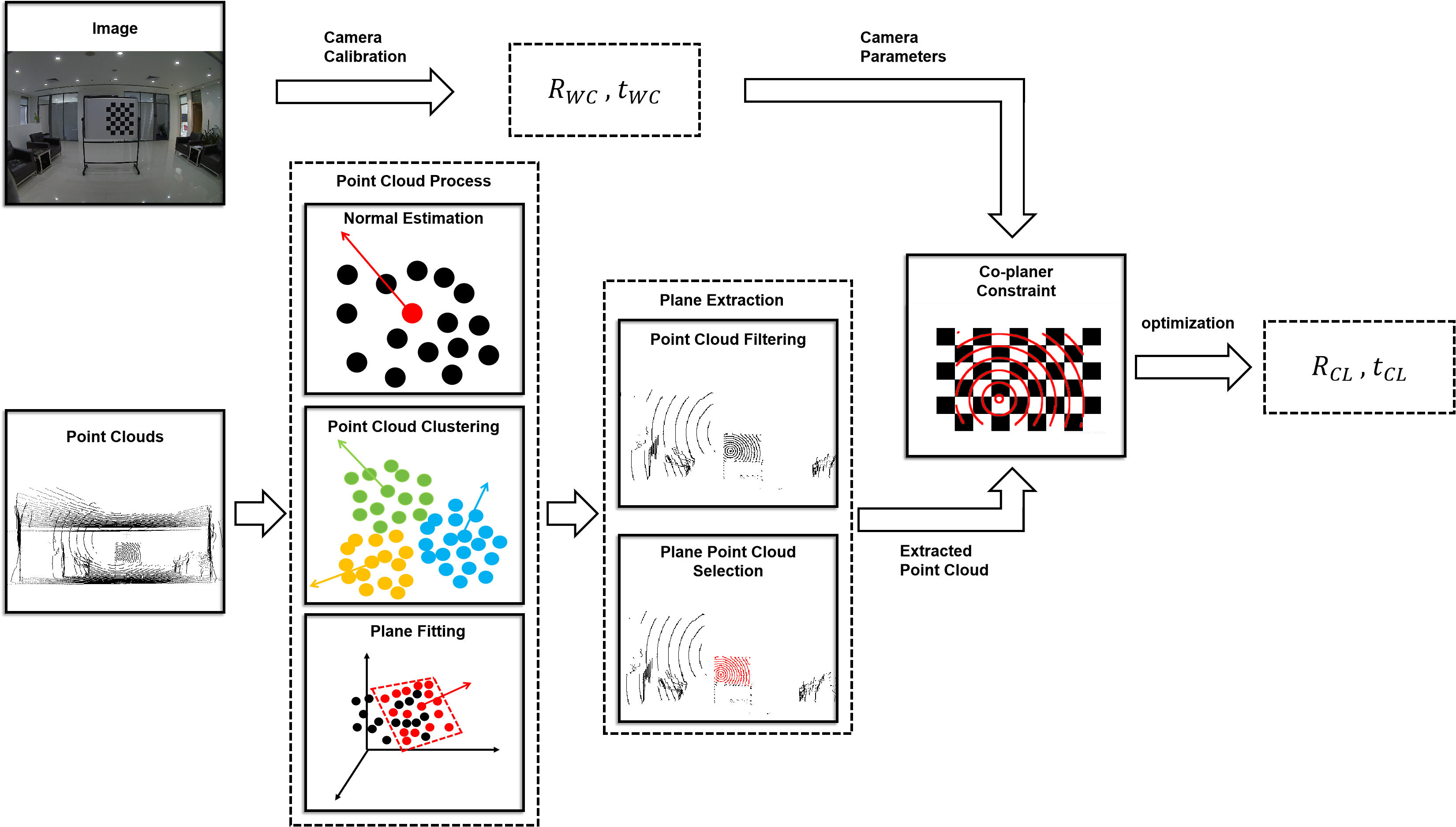}
    \caption{Pipeline of proposed calibration method}
    \label{fig:calibration_pipeline}
\end{figure*}

Fig.\ref{fig:calibration_pipeline} shows the pipeline of the calibration process. After inputting the data, first step is to calibrate the parameter of camera (Section \ref{sec:m1}), then using a proposed novel algorithm to filter point clouds (Section \ref{sec:m2}). Next is to extract the plane point cloud (Section \ref{sec:m2}). The last step is to establish the co-planar constraint and compute the extrinsic parameters by optimizing the cost function of co-planar constraint (Section \ref{sec:m4}).

\section{METHODOLOGY}\label{sec:methodology}
The first step of our proposed method is camera calibration. Following the completion of camera calibration, conventional approaches typically proceed to identify correspondence points between LiDAR and camera, followed by point cloud registration to achieve extrinsic calibration. This process is often challenging and complex. In contrast to these methods, our approach offers superiority by exclusively processing LiDAR point clouds throughout the calibration process. There is no need to register for 2D-to-3D correspondence points between LiDAR and camera or perform point cloud registration. This process is realized through a novel plane point cloud extraction method and the co-planar constraint optimization based on this method.
\begin{algorithm}[!h]
    \caption{Extrinsic Calibration}
    \label{alg:1}
    \renewcommand{\algorithmicrequire}{\textbf{Require:}}
    \renewcommand{\algorithmicensure}{\textbf{Output:}} 
    \begin{algorithmic}[1]
        \REQUIRE Camera images, LiDAR Point cloud \(P_{\text{L}}^i\)  
        \ENSURE \(R_{\text{CL}}\), \(t_{\text{CL}}\)    
\item \(i\)  denotes  \(i\)-th point, \(j\) denotes \(j\)-th point cluster
\item \(k\)  denotes  \(k\)-th plane, \(h\) denotes \(h\)-th filtered plane 
\STATE  From camera calibration get \(R_{\text{WC}}\), \(t_{\text{WC}}\)
\STATE  Calculate threshold \(l\) using Eq.\ref{eq_l}
\FOR{each \(P_{\text{L}}^i\)}
    \STATE Normal vector \(N_{\text{L}}^i\) = PCA(\(P_{\text{L}}^i\))
\ENDFOR
\STATE \(C[j]\) = DBSCAN(\(P_{\text{L}}^i\), \(N_{\text{L}}^i\))  
\FOR{each \(C[j]\)}
    \STATE Calculate \(N_{\text{C}}^j\) using Eq.\ref{eq_nci}
\ENDFOR
\STATE \(Plane[k]\) = RANSAC(\(C_j\), \(N_{\text{C}}^j\))
\STATE Set threshold \(\theta\)
\FOR{each \(Plane[k]\)}
    \STATE Calculate \( \alpha_k\) using Eq.\ref{eq_ai}
\IF{\(\alpha_k > \theta\)}
    \STATE Filter out \(Plane[k]\)
\ENDIF
\IF{\(\alpha_k \leq \theta\)}
    \STATE \(Plane\_filtered[h]\) = \(Plane[k]\)
\ENDIF
\ENDFOR
\FOR{each \(Plane\_filtered[h]\)}
    \STATE Calculate \(d_{\text{\(h\)}}\) using Eq.\ref{eq_di}
    \IF{closest(\(d_{\text{\(h\)}}\), \(l\)) \&\& max(\( \rho_{plane} \)) }
        \STATE Extract \(P_{\text{L}}^i\) from \(Plane\_filtered[h]\)
    \ENDIF
\ENDFOR
\STATE \(R_{\text{CL}}\), \(t_{\text{CL}}\) = Optimization(\(P_{\text{L}}^i\), \(R_{\text{WC}}\), \(t_{\text{WC}}\)) using Eq.\ref{deqn_ex5}
        \STATE \textbf{Return} \(R_{\text{CL}}\), \(t_{\text{CL}}\)
    \end{algorithmic}
\end{algorithm}

\subsection{Camera Calibration}\label{sec:m1}
The first step of the proposed method is to calibrate the camera using a Checkerboard. In the field of camera calibration, method using checkerboard as calibration target has been widely adopted. Methods like \cite{zhang2004extrinsic}, \cite{hu2016extrinsic}, \cite{zhouautomatic}, \cite{miyagawa2010simple}, \cite{chen2016geometry} are capable of calibrating the camera using only few images of a chessboard, obtaining the extrinsic parameters \( R_{\text{WC}} \) and \( t_{\text{WC}} \) from the camera coordinate system \( O_{\text{C}} \) to the checkerboard coordinate system \( O_{\text{W}} \). For a point \( P_{\text{C}}^i \) in \( O_{\text{C}} \), it can be projected onto \( O_{\text{W}} \) using the following formula:
\begin{equation}
\label{deqn_ex1}
P_{\text{W}}^i = R_{\text{WC}} \cdot P_{\text{C}}^i + t_{\text{WC}}
\end{equation}

\begin{figure}
    \centering
    \includegraphics[width=0.9\linewidth]{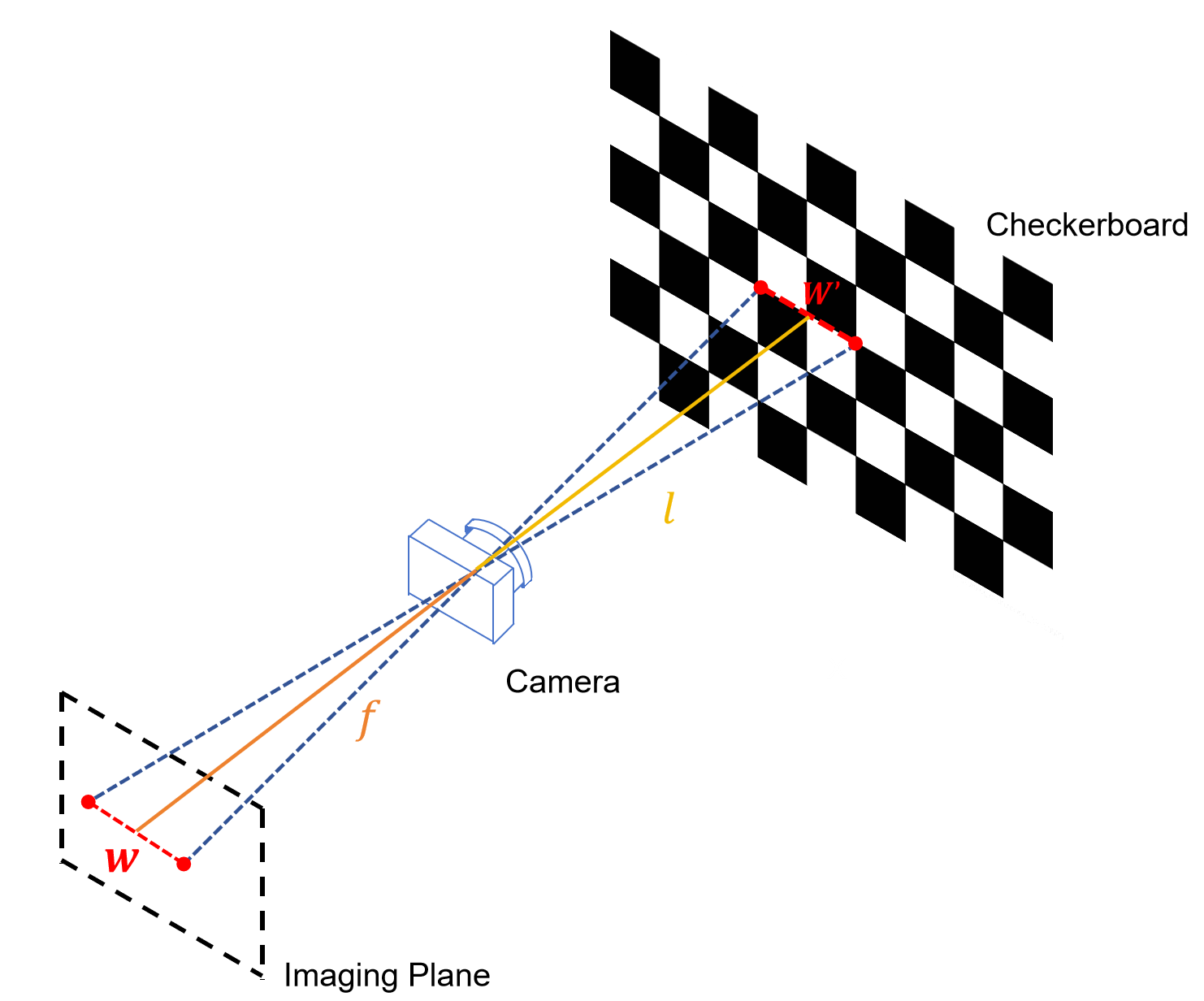}
    \caption{Illustration of the camera pinhole model. The distance between the checkerboard and the camera can be estimated by employing the pinhole model of the camera and the camera parameters}
    \label{fig:distance}
\end{figure}

After the completion of the camera calibration process, an approximation of the distance between the checkerboard and the camera can be achieved through the employment of the pinhole camera model. As shown in Fig. \ref{fig:distance}, let \(f\) denote the focal length of the camera, \(l\) the distance from the checkerboard to the camera, \(w'\) the real-world distance between two distinct corner points on the checkerboard, and \(w\) the corresponding corner point distance within the image. According to the principles of similar triangles, this relationship can be succinctly represented as follows:
\begin{equation}
\label{deqn_ex1}
\frac{f}{w} = \frac{l}{w'} 
\end{equation}
From the camera calibration, we can obtain the value of \(f\), while \(w\) can be ascertained by detecting and calculating the pixel distance between corner points during the camera calibration process. The correspondence \(w'\) can be obtained because we know the size of the checkerboard. Thus, we can obtain the distance from the checkerboard to the camera through the following formula:
\begin{equation}
\label{eq_l}
l = \frac{w \cdot f}{w'}
\end{equation}

This distance is further utilized as a discriminant criterion in the selection process of point cloud extraction in Section \ref{sec:m2}.

\subsection{LiDAR Point Cloud Processing}\label{sec:m2}
As mentioned in Section \ref{sec:methodology}, we need to extract corresponding points of the checkerboard from the LiDAR point cloud to establish co-planar constraints, and then optimize to obtain the extrinsic parameters. To achieve this,
we propose a novel point cloud extraction method based on clustering algorithms, which effectively extracts point clouds belonging to the Checkerboard in complex calibration environments, ensuring the algorithm's effectiveness across different scenarios. Additionally, the proposed approach solely relies on LiDAR point clouds, thereby circumventing the need for corresponding point registration between the camera and LiDAR. This method consists of three main steps: point cloud clustering, point cloud filtering, and plane correspondence point extraction.

Firstly, we perform point cloud clustering, aiming to categorize point clouds into different clusters based on their orientations, facilitating subsequent point cloud filtering. In this process, we initially compute the normals of the point cloud using Principal Component Analysis (\text{PCA}), a simple and efficient method commonly utilized for normal estimation in point clouds. Specifically, for a point \(P_{\text{L}}^i\) within the LiDAR coordinate system, its normal \(N_{\text{L}}^i\) is calculated.

Subsequently, the point cloud is clustered based on the normal directions. When the LiDAR scans objects, there are significant density variations between point clouds on different object surfaces\cite{9766181}. Point clouds within a certain density range are more likely to have similar normal directions. Studies \cite{schubert2017dbscan} have shown that the clustering algorithm Density-Based Spatial Clustering of Applications with Noise (DBSCAN), compared to other clustering methods, often achieves faster clustering speeds and better clustering results when clustering objects with density variations. Therefore, we adopt the DBSCAN method, known for its proficiency in clustering objects based on density differences, to cluster the point cloud.

The DBSCAN algorithm groups the point cloud \(P_{\text{L}}^i\) into clusters based on the normals \(N_{\text{L}}^i\), grouping points with similar normal directions into the same cluster. After clustering, the space point cloud is segmented into different clusters \(C_1, C_2, C_3, \ldots\) with similar orientations. For each cluster, the most representative normal \(N_{\text{C}}^i\) is identified by following equation:
\begin{equation}
\label{eq_nci}
N_{\text{C}}^i = \frac{\sum_{i=1}^{n} W_i \cdot N_{\text{L}}^i}{\|\sum_{i=1}^{n} W_i \cdot N_{\text{L}}^i\|}
\end{equation}
where \(n\) is the total number of point clouds in the cluster, \(W_i\) is the weight of the \(i\)-th point cloud determined based on the density of the point cloud, \(N_{\text{L}}^i\) is the normal of the \(i\)-th point cloud.

The second step involves point cloud filtering, where the Random Sample Consensus (RANSAC) method is utilized for plane fitting owing to its robustness in handling outliers and its efficacy in accurately estimating parameters from noisy data\cite{9393166}. This choice is particularly suitable for identifying the maximum density plane within each cluster of point clouds. Subsequently, only the point clouds forming the maximum density plane are retained in each cluster. Then, the point cloud can be effectively filtered based on the orientation information of the point cloud planes. Considering that the plane of the checkerboard is likely to be parallel to or at a certain angle to the \(xoy\) plane—taking into account the possible tilted placement of the checkerboard—a distinctive characteristic of such planes is that their normal direction will be parallel to the \(z\)-axis or at a certain angle to the \(z\)-axis. Therefore, we have set a threshold \(\theta\) to filter based on the normal direction of each point cloud cluster. We define the angle \(\alpha_i\) as the angle between the normal \(N_{\text{C}}^i\) and the \(z\)-axis. This angle can be calculated using the formula:
\begin{equation}
\label{eq_ai}
\alpha_i = \cos^{-1}\left(\frac{N_{\text{C}}^i \cdot z}{\|N_{\text{C}}^i\| \|z\|}\right)
\end{equation}

where \(N_{\text{C}}^i\) is the representative normal of cluster \(C_i\), \(z\) is the unit vector of the \(z\)-axis (i.e., \(z = [0, 0, 1]^T\)), \(\cdot\) denotes the dot product, and \(\| \|\) denotes the magnitude of the vector. The angle \(\alpha_i\) is calculated by the arc-cosine of the dot product of \(N_{\text{C}}^i\) and \(z\), divided by the product of their magnitudes.

\begin{figure}
    \centering
    \includegraphics[width=1\linewidth]{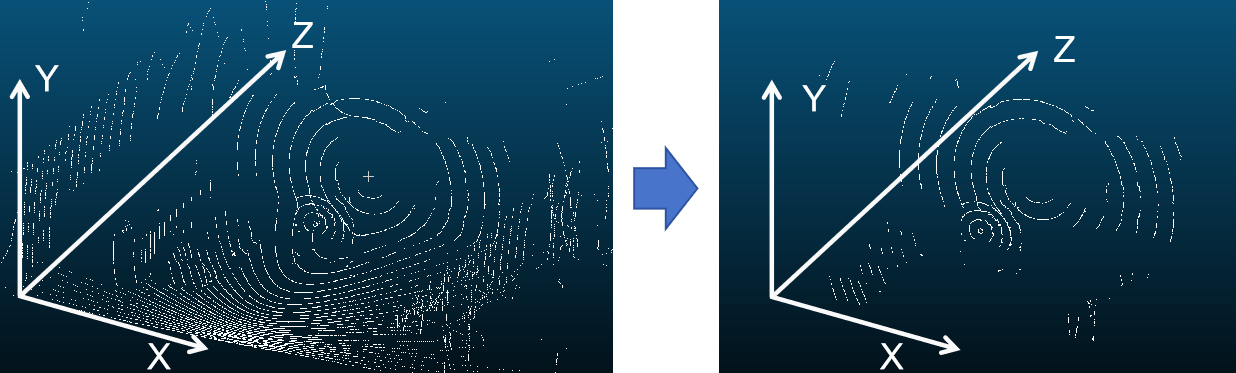}
    \caption{Most of the irrelevant point clouds in the left figure are filtered out, leaving only the point clouds mainly face to the \(xoy\) plane in the right figure.}
    \label{fig:filtering}
\end{figure}

Subsequently, by setting a threshold \(\theta\), we can determine whether this angle meets the requirement. If the angle \(\alpha_i\) between a cluster's normal and the \(z\)-axis is greater than \(\theta\), i.e., \(\alpha_i > \theta\), then the point cloud belonging to that cluster will be filtered out, Fig. \ref{fig:filtering} shows the comparison of point cloud before and after filtering.

Final step is plane point cloud extraction, given the complexity of the actual calibration environment, it is possible that after point cloud filtering, multiple planes still exist. It is necessary to detect the planes belonging to the checkerboard among these multiple planes. This can be achieved by leveraging the prior information obtained from Section \ref{sec:m1}. Specifically, we have computed an approximate distance \( l \) between the checkerboard and the camera. For the general LiDAR-camera systems installed in close proximity, this \( l \) can be directly utilized as an approximate distance threshold. If there is a significant disparity in the installation distance between the LiDAR and the camera, we can also manually measure and adjust the value of \( l \). We aim to identify the filtered plane cluster \( C_i \) whose distance \( d_i \) to the LiDAR is closest to \( l \).
The point clouds in cluster $C_i$ can be represented as a $3 \times n$ matrix $\mathbf{A}_i = [x_1, y_1, z_1; x_2, y_2, z_2; ...; x_n, y_n, z_n]^T$. Summing across each row of the matrix, we can obtain a $3 \times 1$ matrix $\mathbf{S}_i = [\sum_{i=1}^{n} x_i, \sum_{i=1}^{n} y_i, \sum_{i=1}^{n} z_i]^T$. The distance $d_i$ can be obtained by the following equation:

\begin{equation}
\label{eq_di}
\begin{aligned}
d_{\text{i}} = \frac{1}{n} \cdot \sqrt{\mathbf{S}_i^T \cdot \mathbf{S}_i}
\end{aligned}
\end{equation}

    

\begin{figure}
    \centering
    \includegraphics[width=1\linewidth]{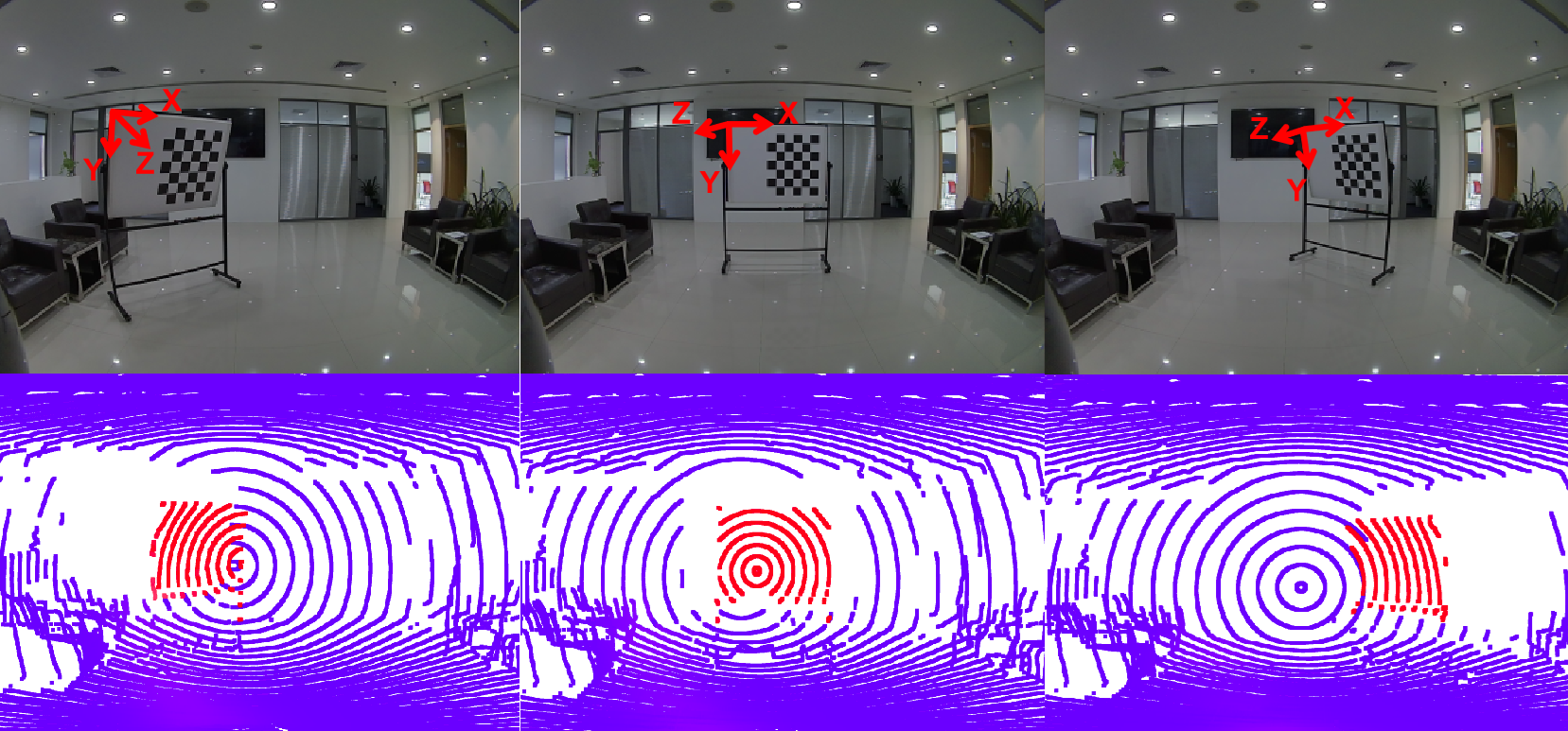}
    \caption{Plane point cloud extraction in different scenario.}
    \label{fig:extraction}
\end{figure}

Where \(\mathbf{S}_i^T\) is the transpose of \(\mathbf{S}_i\). 
 In complex environments, it is possible to encounter multiple \( d_i \) values close to or even equal to \( l \). To address this, we introduce the point cloud density \( \rho \) as another assessment, where the value of \( \rho \) equals the number of point clouds in the cluster. In our calibration environment requirements, we aim to use relatively large checkerboard, ensuring that the point cloud density \( \rho \) on the checkerboard is maximal. Through this process, we can ultimately determine the point cloud in the LiDAR data that belongs to the checkerboard by finding the cluster with the closest \( d_i \) value to \( l \) as well as the largest point cloud density \( \rho \) in the candidate cluster. Finally, we extract the point cloud in the selected cluster, denoted as \( P_{\text{L}}^i \). These are the corresponding points of checkerboard in LiDAR point clouds. Fig. \ref{fig:extraction} illustrates the precise extraction of corresponding points in different calibration scenarios.

\subsection{Co-planar Constraint and Extrinsic Calculation}\label{sec:m4}
For the point cloud \(P_{\text{L}}^i\) extracted in the previous step, if we know the extrinsic parameters \(R_{CL}\) and \(t_{CL}\), we can project it from the LiDAR coordinate system to the camera coordinate system, yielding the point cloud \(P_{\text{C}}^i\):
\begin{equation}
\label{deqn_ex1}
P_{\text{C}}^i = R_{CL} \cdot P_{\text{L}}^i + t_{CL}
\end{equation}

Utilizing the extrinsic parameters of the camera to the checkerboard coordinate system, \(R_{WC}\) and \(t_{WC}\), we can further project \(P_{\text{C}}^i\) from the camera coordinate system to the checkerboard coordinate system, resulting in the point cloud \(P_{\text{W}}^i\):
\begin{equation}
\label{deqn_ex2}
P_{\text{W}}^i = R_{WC} \cdot (R_{CL} \cdot P_{\text{C}}^i + t_{CL}) + t_{WC}
\end{equation}
Since \(P_{\text{W}}^i\) belongs to the \(xoy\) plane of the checkerboard, it should lie on the plane of the checkerboard during scanning, indicating that the \(z\)-axis coordinates of these points should be zero. Thus, we can establish the constraint equation:
\begin{equation}
\label{deqn_ex3}
z(P_{\text{W}}^i) = z(R_{WC} \cdot (R_{CL} \cdot P_{\text{L}}^i + t_{CL}) + t_{WC}) = 0
\end{equation}
We aim to minimize the \(z\)-axis coordinates of the checkerboard point \(P_{\text{W}}^i\) to solve for the unknown extrinsic parameters \(R_{CL}\) and \(t_{CL}\). This can be achieved by minimizing the sum of squared errors:

\begin{equation}
\label{deqn_ex4}
\text{minimize} \sum_{i=1}^{N} (z(P_{\text{W}}^i))^2
\end{equation}
Here, \(N\) is the number of extracted LiDAR points, and \(z(P_{\text{W}}^i)\) represents the \(i\)-th point's z value in the checkerboard coordinate system. Substituting into Equation \ref{deqn_ex1}, it can write as:
\begin{equation}
\label{deqn_ex5}
\argmin_{R_{CL},t_{CL}} \sum_{i=1}^{N} \left| z(R_{WC} \cdot (R_{CL}\cdot P_{\text{L}}^i + t_{CL}) + t_{WC}) \right|^2
\end{equation}
By solving this optimization problem, we can obtain the desired extrinsic parameters \(R_{CL}\) and $t_{CL}$, thereby achieving the calibration between the LiDAR and the camera. This optimization process relies solely on the constraints provided by the LiDAR point cloud. Moreover, since each point on the plane serves as a constraint, we have established a sufficient number of constraints in one input frame to optimize accurate values for the extrinsic parameters. After several iterations of optimization, we can ultimately determine the precise extrinsic parameters $R_{CL}$ and $t_{CL}$ between the LiDAR and the camera. 

The whole calibration algorithm is summarized in Algorithm ~\ref{alg:1}.
It is noteworthy that, from processing the LiDAR point cloud to obtaining the extrinsic parameters between the camera and the LiDAR through optimization, we exclusively handle the LiDAR point cloud and do not perform any corresponding point registration operations. This eliminates the need for registering correspondence between points and reduces potential errors caused by correspondence mismatches.

\section{EXPERIMENT}\label{sec:experiment}
\begin{figure}
    \centering
    \includegraphics[width=1\linewidth]{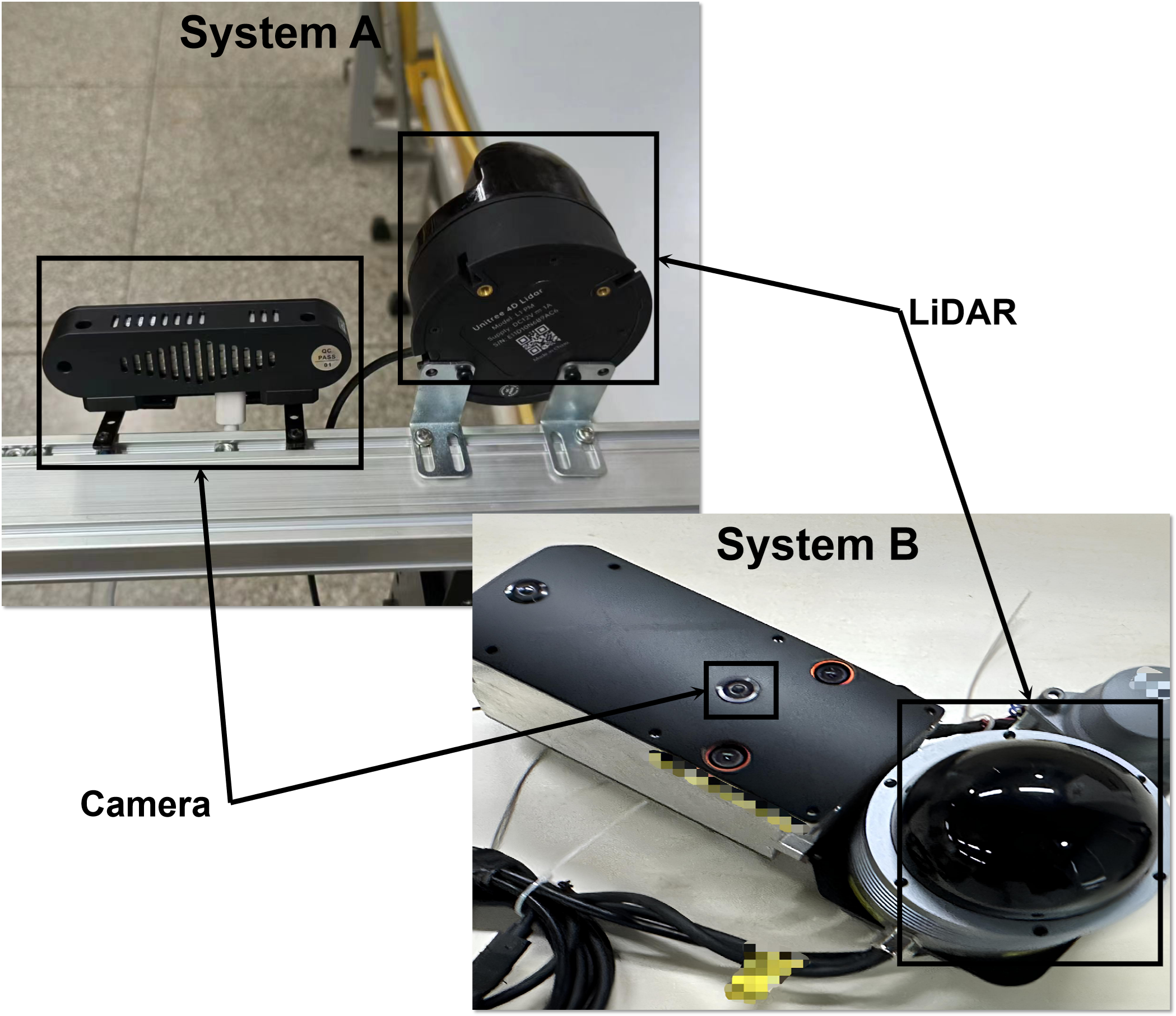}
    \caption{System A consists of a Gemini Pro camera and a Unitree L1 LiDAR; System B consists of an X1 camera and a Robosense LiDAR.}
    \label{fig:device}
\end{figure}

We conducted a series of experiments, encompassing both simulated and real-world data, to evaluate the performance of our proposed method. In Section \ref{simuexp}, we assess the method's effectiveness by calculating calibration accuracy, calibration running time and the translation and rotation errors of extrinsic, leveraging the availability of ground truth data in the simulated environment. In Section \ref{realexp} experiments, the proposed method has been validated on real-world data collected by two LiDAR-Camera systems A and B shown in Fig.\ref{fig:device}. We evaluate the algorithm's performance by visualizing the extraction process and re-projecting the LiDAR point cloud onto camera images. We will conduct comparative experiments, comparing the results generated by the proposed method in this paper with the results produced by the four State of the Art(SOTA) methods shown in Table~\ref{compare_table}. All experiments were conducted on a regular laptop with Intel Core i5-8250U CPU and GPU of NVIDIA GeForce MX150.
\begin{table*}
\centering
\caption{PROPERTIES OF THE COMPARISON METHODS}
\label{compare_table}
\begin{tabular}{cccc}
\toprule
Method & Method Type & Calibration Target & Minimum Calibration Frame \\
\midrule
Geiger\cite{geiger2012automatic} & Fully Automatic & At least four checkerboards & 1 \\
Park\cite{park2014calibration} & Manual & At least three whiteboards & 12 \\
Pusztai\cite{pusztai2017accurate} & Semi-automatic & One box with three perpendicular sides & 6 \\
Matlab Toolbox\cite{zhouautomatic} & Semi-automatic & One checkerboard & 6 \\
\textbf{Proposed} & \textbf{Fully Automatic} & \textbf{One checkerboard} & \textbf{2} \\
\bottomrule
\end{tabular}
\end{table*}

\begin{table*}[htbp]
  \centering
  \caption{THE MEAN AND STANDARD DEVIATION OF CALIBRATION ERRORS FOR THE PROPOSED METHOD AND THE COMPARISON METHODS ACROSS THREE LIDAR-CAMERA SYSTEMS}
    \begin{tabular}{cccccccc}
    \toprule
    \multirow{2}{*}{Method} & \multirow{2}{*}{Metric} & \multicolumn{3}{c}{Rotation Error(°)} & \multicolumn{3}{c}{Translation Error(m)} \\
    \cmidrule(lr){3-5} \cmidrule(lr){6-8} 
    & & Roll & Pitch & Yaw & X & Y & Z \\
    \midrule
    \multirow{2}{*}{\centering Geiger} & Avg & 0.4001 & 0.4067 & 0.4503 & 0.0853 & 0.0903 & 0.0970 \\
          & Std & 0.0258 & 0.0338 & 0.0141 & 0.0082 & 0.0053 & 0.0041 \\
    \addlinespace[0.5em]
    \multirow{2}{*}{\centering Park} & Avg & 0.3633 & 0.3233 & 0.3933 & 0.0892 & 0.1002 & 0.0933 \\
          & Std & 0.0350 & 0.0661 & 0.0573 & 0.0070 & 0.0121 & 0.0050 \\
    \addlinespace[0.5em]
    \multirow{2}{*}{\centering Pusztai} & Avg & 0.2933 & 0.3133 & 0.2833 & 0.0727 & 0.0760 & 0.0733 \\
          & Std & 0.0140 & 0.0381 & 0.0323 & 0.0037 & 0.0060 & 0.0062 \\
    \addlinespace[0.5em]
    \multirow{2}{*}{\centering Matlab Toolbox} & Avg & 0.2467 & 0.2467 & 0.3233 & 0.0626 & 0.0673 & 0.0623 \\
          & Std & 0.0281 & 0.0170 & 0.0254 & 0.0033 & 0.0091 & 0.0051 \\
    \addlinespace[0.5em]
    \multirow{2}{*}{\centering \textbf{Proposed}} & \textbf{Avg} & \textbf{0.0171} & \textbf{0.0430} & \textbf{0.0662} & \textbf{0.0173} & \textbf{0.0147} & \textbf{0.0103} \\
          & \textbf{Std} & \textbf{0.0071} & \textbf{0.0043} & \textbf{0.0051} & \textbf{0.0004} & \textbf{0.0013} & \textbf{0.0002} \\
    \bottomrule
    \end{tabular}%
  \label{error_compare}%
\end{table*}

\begin{table*}[htbp]
  \centering
  \caption{COMPARED WITH THE RESULTS OF SOTA CALIBRATION METHODS IN KITTI DATASET.}
    \begin{tabular}{ccccccccc}
    \toprule
    Method & \multicolumn{1}{c}{$Error_{\text{X}}^\text{m}$} & \multicolumn{1}{c}{$Error_{\text{Y}}^\text{m}$} & \multicolumn{1}{c}{$Error_{\text{Z}}^\text{m}$} & \multicolumn{1}{c}{$Error_{\text{Mean}}^\text{m}$} & \multicolumn{1}{c}{$Error_{\text{Roll}}^\circ$} & \multicolumn{1}{c}{$Error_{\text{Pitch}}^\circ$} & \multicolumn{1}{c}{$Error_{\text{Yaw}}^\circ$} & \multicolumn{1}{c}{$Error_{\text{Mean}}^\circ$} \\
    \midrule
    \multicolumn{9}{c}{KITTI$_{1}$} \\
    Zhu\cite{zhu2020online} & 0.073 & 0.085 & 0.083 & 0.080 & 0.388 & 0.278 & 0.128 & 0.264 \\
    Ma\cite{ma2021crlf} & 0.080 & 0.063 & 0.079 & 0.074 & 0.236 & 0.542 & 0.392 & 0.390 \\
    Calibnet\cite{pei2023calibnet} & 0.091 & 0.103 & 0.077 & 0.090 & 0.489 & 0.321 & 0.236 & 0.348 \\
    ATOP\cite{sun2022atop} & 0.064 & 0.063 & 0.057 & 0.061 & 0.258 & 0.214 & 0.324 & 0.265 \\
    \textbf{Proposed} & \textbf{0.011} & \textbf{0.018} & \textbf{0.016} & \textbf{0.015} & \textbf{0.172} & \textbf{0.113} & \textbf{0.132} & \textbf{0.139} \\
    \multicolumn{9}{c}{KITTI$_{2}$} \\
    Zhu\cite{zhu2020online} & 0.063 & 0.066 & 0.071 & 0.067 & 0.228 & 0.211 & 0.158 & 0.199 \\
    Ma\cite{ma2021crlf} & 0.072 & 0.055 & 0.069 & 0.065 & 0.256 & 0.343 & 0.212 & 0.270 \\
    Calibnet\cite{pei2023calibnet} & 0.053 & 0.077 & 0.076 & 0.068 & 0.291 & 0.121 & 0.251 & 0.221 \\
    ATOP\cite{sun2022atop} & 0.051 & 0.064 & 0.053 & 0.056 & 0.398 & 0.241 & 0.312 & 0.317 \\
    \textbf{Proposed} & \textbf{0.014} & \textbf{0.011} & \textbf{0.009} & \textbf{0.011} & \textbf{0.122} & \textbf{0.083} & \textbf{0.092} & \textbf{0.099} \\
    \bottomrule
    \end{tabular}%
  \label{tab:kitti}%
\end{table*}%

\subsection{Simulation Experiment}\label{simuexp}


In the simulation experiment, a calibration scenario was created in Gazebo Robot Simulation Environment (Gazebo) to validate both the calibration accuracy and the generalizability of our method across diverse LiDAR-camera systems. We performed calibration on three distinct setups, each employing different combinations of cameras and LiDARs, with variations in the displacement between the camera and LiDAR for each system. Four comparative methods listed in Table~\ref{compare_table} were employed for comparison alongside the proposed method in this paper. For the methods used for comparison, we use the same simulation environment and adhered to their default calibration procedures. Data were collected using three sets of equipment, and both the proposed method and comparative methods were employed to calibrate the camera and LiDAR systems using same collecting data. We employ two primary metrics: rotation error $Error_{\text{Roll,Pitch,Yaw}}^\circ$ and translation error $Error_{\text{X,Y,Z}}^\text{m}$:
\begin{equation}
\label{error1}
Error_{\text{Roll,Pitch,Yaw}}^\circ = \arccos\left(\frac{\text{trace}(\mathbf{R}_{\text{c}} \mathbf{R}_{\text{true}}^T) - 1}{2}\right)
\end{equation}
\begin{equation}
\label{error2}
Error_{\text{X,Y,Z}}^\text{m} = \|\mathbf{t}_{\text{c}} - \mathbf{t}_{\text{true}}\|_2
\end{equation}
where $\mathbf{R}_{\text{c}}$ represents the rotation matrix of the calibration result, $\mathbf{R}_{\text{true}}$ represents the ground truth matrix; $\mathbf{t}_{\text{c}}$ represents the translation vector of the calibration result, $\mathbf{t}_{\text{true}}$ represents the ground truth vector. $\text{trace}(\cdot)$ denotes the trace of the matrix. $\|\cdot\|_2$ is the L2 norm of the vector. To more clearly demonstrate the accuracy of the method, we calculated the specific errors in the XYZ, roll, pitch, and yaw dimensions. 

Table~\ref{error_compare} presents the mean and standard deviation of these errors for these methods across the three different LiDAR-camera systems. It can be observed that the proposed method exhibits rotation errors of 0.017, 0.043, and 0.066 in terms of roll, pitch, and yaw respectively, and translation errors of 0.0173, 0.0147, and 0.0103 meters in the X, Y, and Z directions respectively. The small errors imply that the results obtained by the proposed method are closest to the ground truth, indicating superior calibration accuracy compared to the comparative methods. Additionally, the proposed method demonstrates the smallest standard deviation across the data, indicating its robustness in accurately calibrating LiDAR-camera systems with different configurations. 
\begin{figure}
    \centering
    \includegraphics[width=1\linewidth]{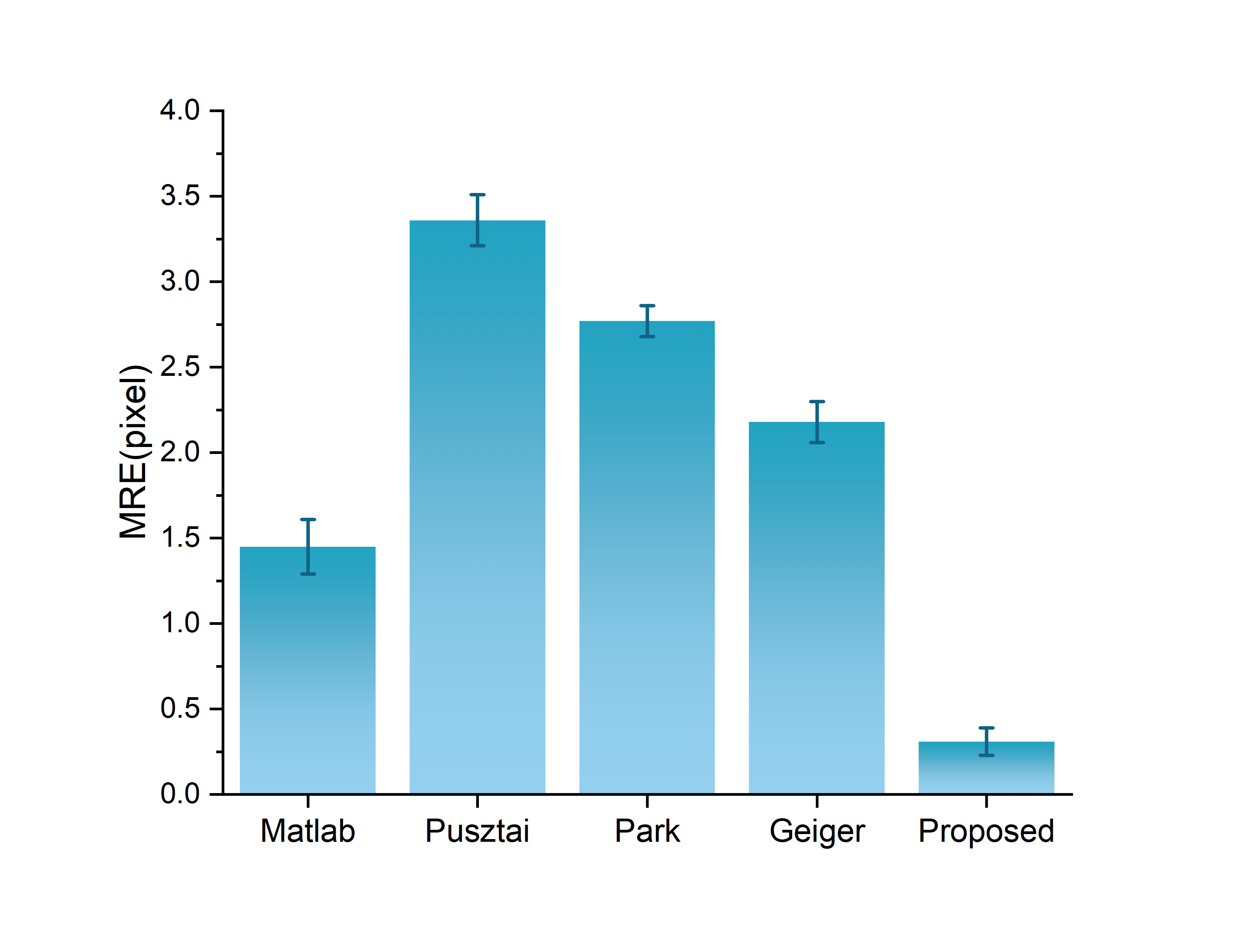}
    \caption{ Mean re-projection error(MRE) for different calibration methods.}
    \label{fig:reproject}
\end{figure}
\renewcommand{\floatpagefraction}{.8}
\begin{figure*}
    \centering
    \includegraphics[width=0.8\linewidth]{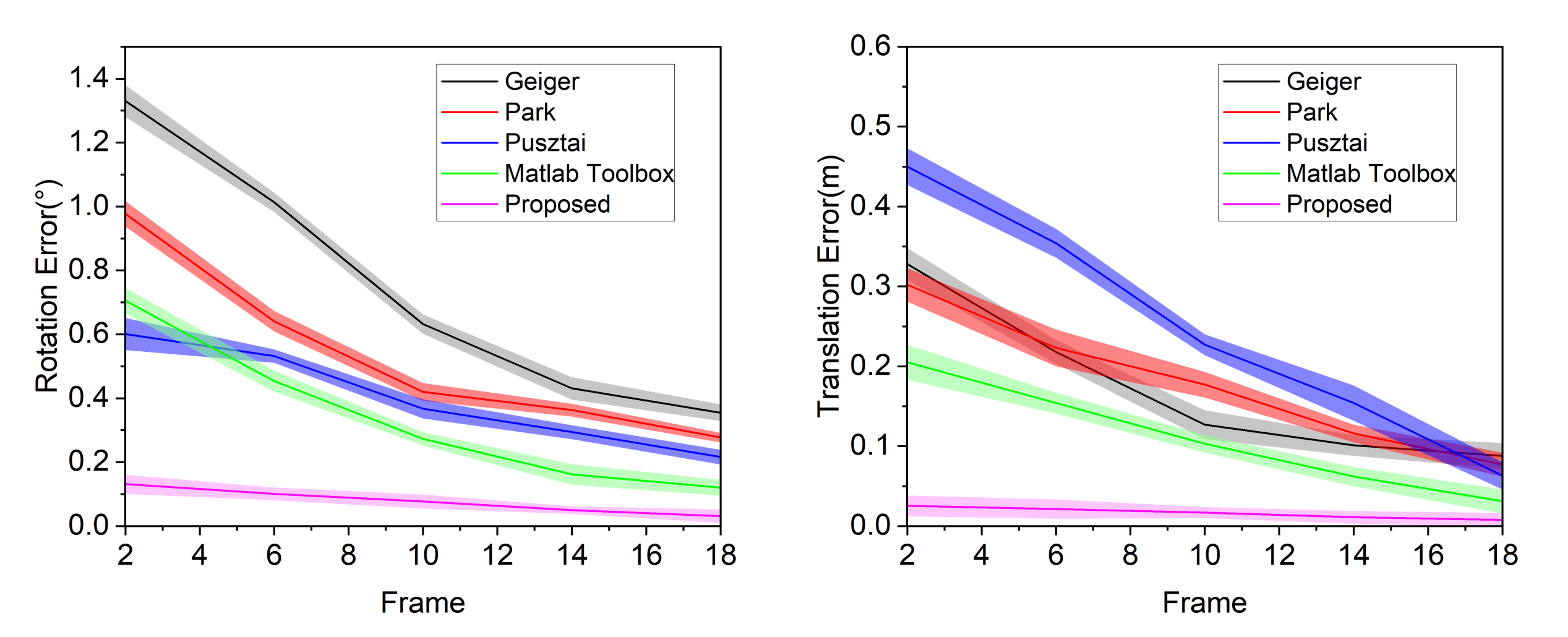}
    \caption{Error band chart of calibration accuracy for different input frames.}
    \label{fig:lineerrorchart}
\end{figure*}
\renewcommand{\floatpagefraction}{.8}
\begin{figure*}
    \centering
    \includegraphics[width=0.8\linewidth]{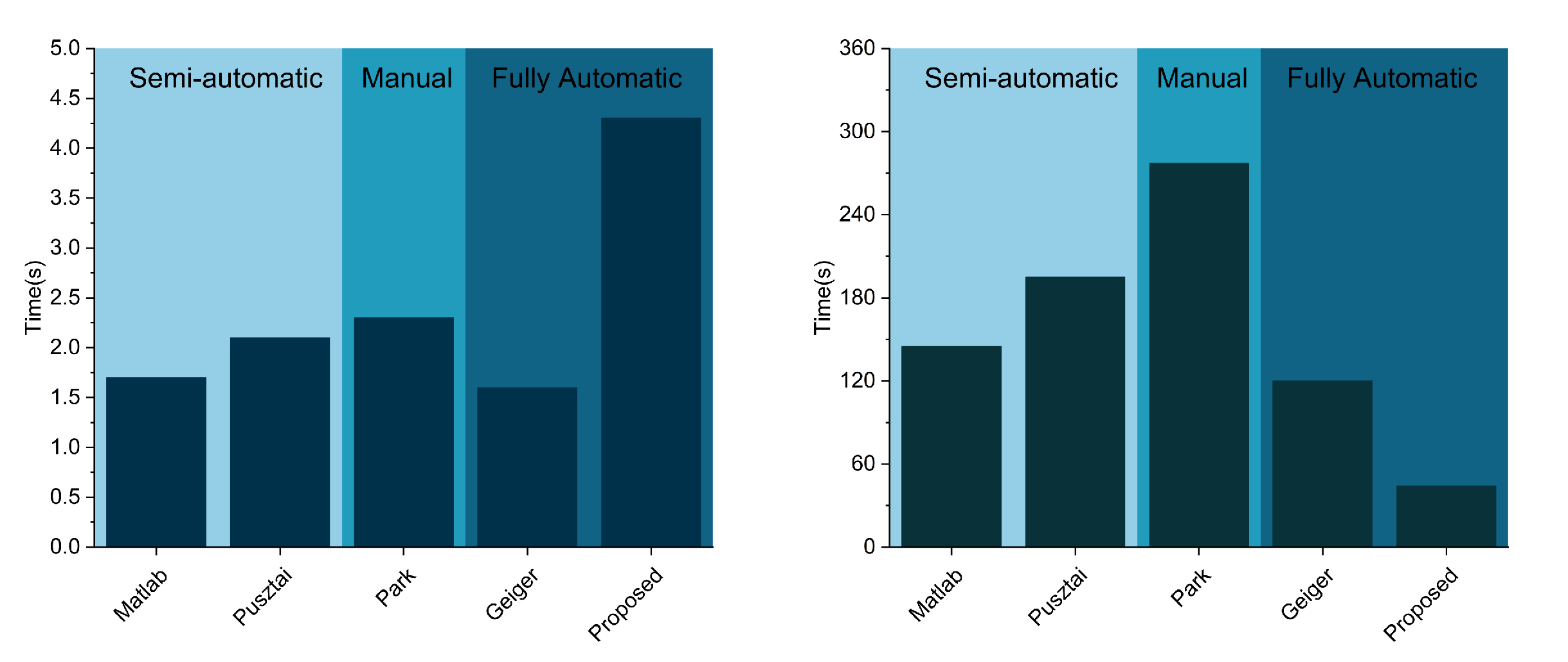}
    \caption{Left is the time consumption for processing each frame; Right is the total calibration time.}
    \label{fig:timecost}
\end{figure*}

After the calibration accuracy experiment, we also computed the mean re-projection error (MRE) for different calibration methods. In the simulated environment, a uniform circular plane was vertically placed in front of the equipment, and all other objects were removed. Data were collected using three equipment, and then, a manual selection of the circular center point cloud was performed. Subsequently, the center point cloud was re-projected onto the camera coordinate system through the extrinsic parameters. The pixel distance between the re-projected center point \( \mathbf\mu' = (u', v') \) and the center of the circular image \( \mathbf\mu = (u, v) \) in the camera was calculated as the re-projection error:
\begin{equation}
\label{reprojerror2}
Error_\text{rep} = \sqrt{(u' - u)^2 + (v' - v)^2}
\end{equation}
Fig.~\ref{fig:reproject} demonstrates that the proposed method achieves the minimum MRE , which is less than 0.5 pixels.

To highlight the advantage of the proposed method in requiring less data, in the subsequent experiments, we will set the input calibration frames as a control variable. Calibration frame is defined as an image synchronized with its corresponding LiDAR point cloud data. This approach allows us to assess and compare the calibration accuracy of both the proposed method and the SOTA methods under various quantities of input calibration frames, thereby illustrating the efficiency of our approach in terms of data utilization.

As observed from Fig.\ref{fig:lineerrorchart}, the proposed method significantly outperforms the comparative methods in calibration accuracy when utilizing a small number of calibration frames. This can be attributed to the proposed method's efficient utilization of point cloud information through the extraction of planar point clouds and the establishment of co-planar constraints. Even with minimal data, the method is capable of constructing strong point cloud constraints. In contrast, correspondence registration based methods generally require a larger volume of input to establish substantial constraints that ensure calibration accuracy. This phenomenon is evident as the accuracy of all methods gradually improves with an increasing number of calibration frames. Methods such as the Geiger and Park exhibit some variability in standard deviation as the number of calibration frames increases. This variability is due to the introduction of more correspondence points by additional calibration frames, which, while increasing the number of correspondences, also leads to a higher incidence of correspondence mismatches, thereby causing fluctuations.

Utilizing a increasing number of calibration frames is beneficial for enhancing calibration accuracy; however, the inclusion of additional input data necessitates extended calibration time. Fig.\ref{fig:timecost} illustrates the per-frame processing time as well as the total calibration time across different methodologies. Fully automated approaches hold a significant advantage in terms of processing time, with all methods, without exception, experiencing increased processing times as the volume of calibration data inputs grows. In terms of processing speed, the method proposed in this paper requires a longer duration to process calibration frame. However, the proposed method achieves the accuracy attainable by SOTA methods with a multitude of input datasets using only a limited number of input frames. Therefore, the overall calibration time still remains ahead. In summary, the method introduced in this paper demands less calibration data and possesses an advantage in calibration speed, making it both efficient and effective.

\subsection{Real-world Data Experiment}\label{realexp}
This real-world data experiment includes re-projecting the LiDAR point clouds into image captured by the camera and evaluating proposed method using KITTI \cite{geiger2013vision} dataset.
\begin{figure*}[ht]
    \centering
    \includegraphics[width=0.8\linewidth]{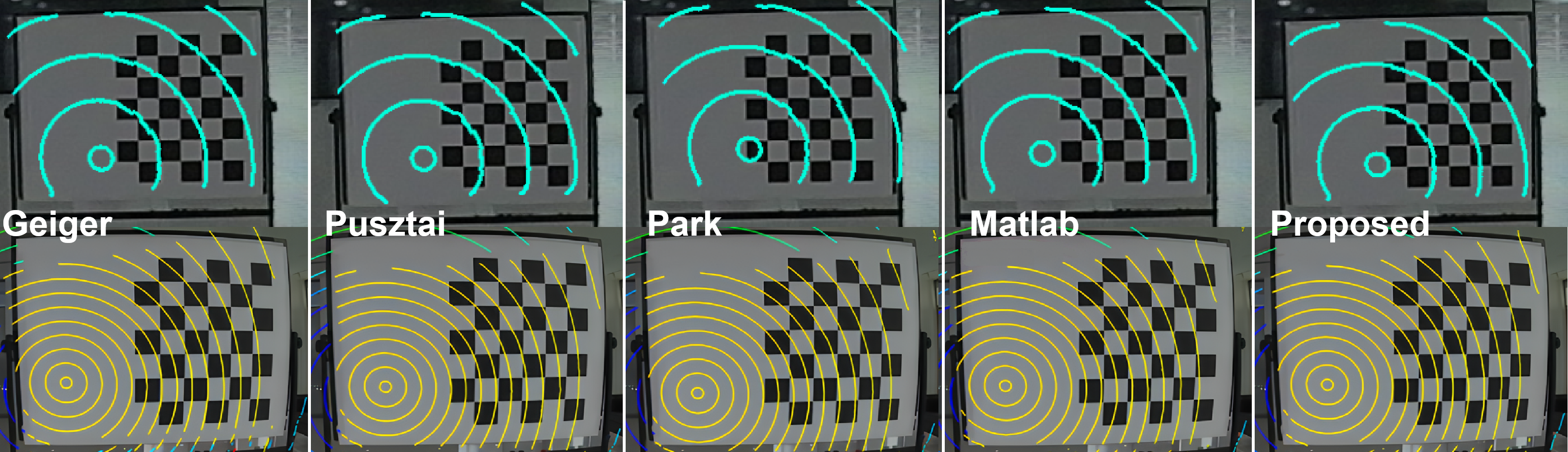}
    \caption{The first row shows the re-projection results of different calibration methods applied to LiDAR-camera System A, while the second row shows the re-projection results of the same methods applied to System B. }
    \label{fig:indoorrep}
\end{figure*}

\begin{figure*}
    \centering
    \includegraphics[width=0.8\linewidth]{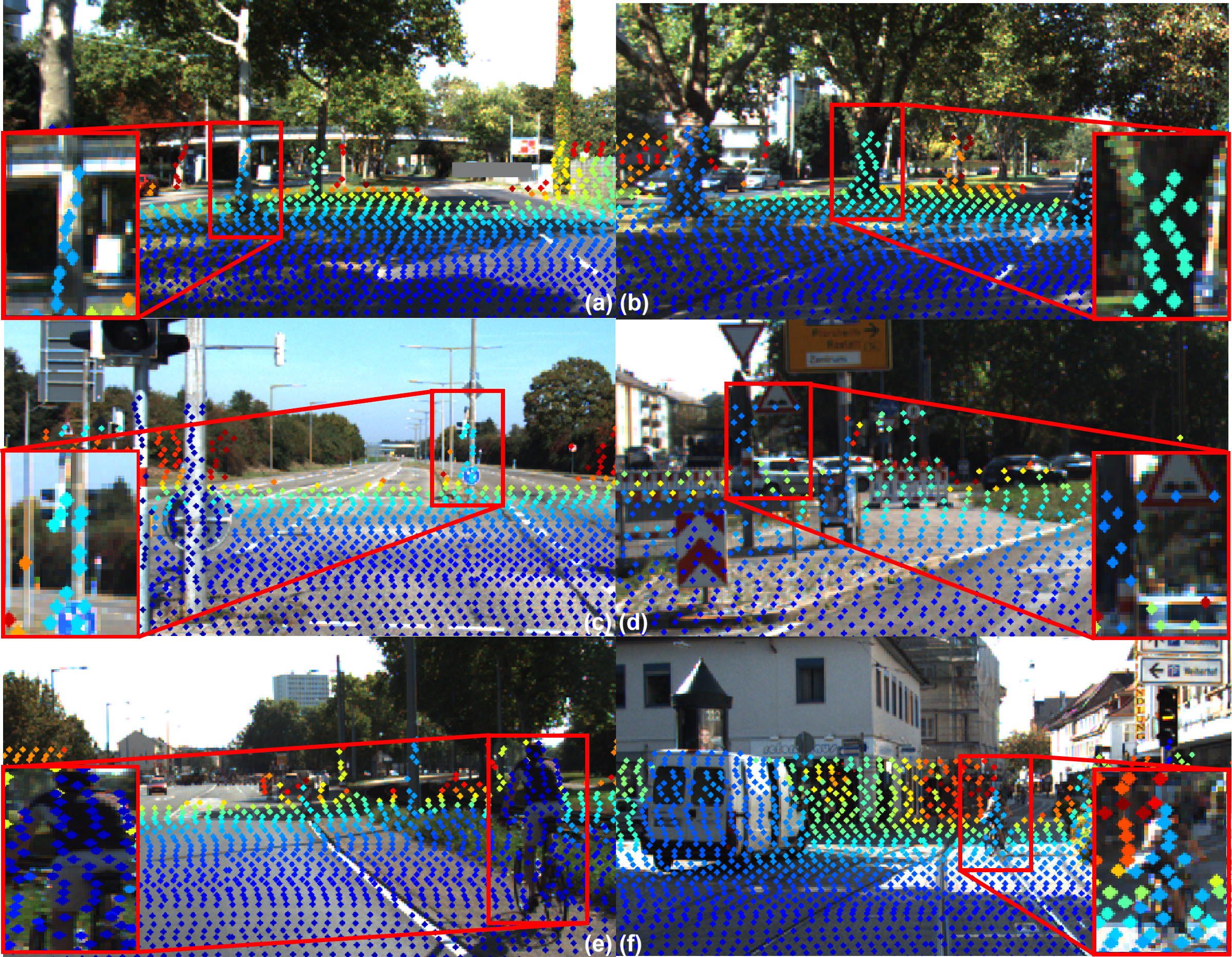}
    \caption{(a)(b), (c)(d), (e)(f) illustrate the re-projection of trees, roadside poles, and cyclist respectively. The point cloud of trees and poles is concentrated on the main body without exceeding the boundaries, and the point cloud projection on the cyclist is uniform in color (depth), indicating precise extrinsic calibration.}
    \label{fig:outdoorrep}
\end{figure*}

Fig.\ref{fig:indoorrep} represents the re-projection outcome, demonstrating the superior performance of proposed method, particularly evident at the edges of the checkerboard. In our method, the laser points belonging to the checkerboard are accurately projected within its boundaries, whereas alternative approaches tend to project some of these points outside the checkerboard, to varying degrees. This observation underscores the higher calibration accuracy and generalizability achieved by proposed method.

As detailed in Section \ref{sec:methodology}, our method relies solely on plane identification when camera parameters are known. Consequently, our approach can be evaluated using the KITTI dataset by manually selecting the appropriate plane, given the availability of camera parameters. In contrast, other calibration methods previously compared necessitate specific calibration targets not found in the KITTI dataset. Accordingly, we selected several SOTA calibration methods \cite{zhu2020online},  \cite{ma2021crlf}, \cite{pei2023calibnet}, \cite{sun2022atop} suitable for the KITTI dataset for comparison.
We selected two sequences from the KITTI dataset that contain the requisite plane for calibration, enabling the calibration of the camera-LiDAR system and the computation of errors relative to ground truth, as presented in Table~\ref{tab:kitti}.

From Table~\ref{tab:kitti}, it is evident that our method consistently achieves outstanding performance on the KITTI dataset, compared to other methods, the calibration accuracy on the two sequence has been improved by up to 64.36\% and 68.77\%, respectively. This means that in outdoor environments, the measurement errors caused by calibration inaccuracies of the LiDAR-camera system at a distance of 20 \(m\) has decreased from a maximum of 14 \(cm\) to 5 \(cm\). Moreover, we utilize our calibration parameters to re-project outdoor LiDAR point clouds, as depicted in Fig.~\ref{fig:outdoorrep}. We selected two trees, two roadside poles, and two cyclists as our observation targets. The color of the point clouds is determined by depth, revealing consistent and evenly distributed colors on the observation targets, aligning well with their contours. This indicates that our method yields highly accurate re-projection results.

\section{CONCLUSION}\label{sec:conclusion}
This paper introduces a novel LiDAR-camera calibration method that is straightforward to implement and does not rely on human intervention. The key innovation of this work lies in circumventing corresponding point registration using in current most calibration methods. This is achieved through an algorithm capable of extracting plane point clouds in complex environments and establishing co-planar constraints. Specifically, the algorithm calculates point cloud normals, fits planes, filters candidate planes using normal filtering, and applies distance and density thresholds to finally extract LiDAR point clouds. Co-planar constraints are then built using the extracted LiDAR point clouds, and the extrinsic parameters between LiDAR and camera are introduced. These parameters are ultimately obtained through optimization. This process circumvents the errors introduced by mismatched corresponding points in registration-based methods and maximizes the utilization of point cloud information, thereby reducing the required calibration input data. Extensive experiments on both simulation and real-world data have demonstrated that this method exhibits high accuracy in extrinsic calibration compared to current methods. The average rotation and translation errors of the calibration results are both less than 0.05° and 0.015m, respectively.

Although the proposed method offers these advantages, there are still areas for improvement and enhancement. Firstly, the algorithm for extracting planar point clouds in the method may struggle to accurately extract the checkerboard plane when there are many planes in the environment that resemble the checkerboard. Additionally, when the checkerboard is placed too far from the LiDAR-camera setup, the sparse point cloud of its plane may lead to insufficiently strict co-planar constraints, thereby affecting the accuracy of calibration parameters. Future work involves further enhancing the effectiveness and robustness of the planar point cloud extraction method in complex scenarios and extending the algorithm to more general scenes. Moreover, efforts will be made to organize the entire work, develop calibration software, and release it as open-source, to serve a wider audience of researchers and industrial practitioners.

\bibliographystyle{IEEEtran}
\bibliography{IEEEabrv,refs}

\vfill

\end{document}